# Financial Distress Prediction For Small And Medium Enterprises Using Machine Learning Techniques

Yuan Gao[1], Biao Jiang, Jietong Zhou

[1]College of Computer Science and Technology, Harbin Engineering University  one@hrbeu.edu.cn

*Abstract*— Financial Distress forecast assumes a fundamental part in the monetary peculiarity. The exact forecast of the number and plausibility of bombing structures goes about as a list of the development and strength of a country's economy. Financial distress expectation is a key test each supporting supplier faces while deciding borrower reliability. The innate darkness of Small and Medium Enterprise businesses entangles credit dynamic interaction, accordingly inflating the cost to back and bringing down the likelihood of getting reserves. Customarily, a few strategies have been introduced for viable FCP. Then again, the arrangement execution, expectation exactness, and data lawfulness aren't adequate for useful applications. Furthermore, many of the created strategies perform well for a portion of the dataset yet are not versatile to various datasets. Consequently, there is a prerequisite to foster a productive expectation model for better order execution and versatility to different datasets. In this review, we propose a calculation for highlight determination order expectation in light of element credits and data source gathering. the current financial distress expectation model generally just purposes the data from financial explanation and disregards the idealness of organization test by and by. hence, we propose a corporate financial distress expectation model that is better following the training and joins the gathering meager head part examination of financial data, corporate administration qualities, and market exchange data with a Relevant vector machine. Exploratory outcomes demonstrate the way that this strategy can further develop the forecast proficiency of financial distress with fewer trademark factors

## I. INTRODUCTION

Because of changes in business sectors and the economy in itself, financial distress expectation is critical. Moneylenders and investors, officials, national banks, inspectors, and supervisors esteem opportune data regarding a company's financial well-being. The capacity to foresee financial distress is vital for the actual organizations, to expand their true capacity, keep up with as well as increment the quantity of current financial backers, and amplify the stock worth. From various examination studies, it has been laid out that serious financial distress sabotages the financial maintainability of enterprises. The discovery of corporate disappointment can advance the financial manageability of enterprises. A few definitions have been credited to financially temperamental organizations, for example, corporate or business disappointment, illiquidity, indebtedness, or insolvency[1][2][3].

The prosperity of nations and the global economy relies heavily on SMEs. However, one of the major challenges faced by SMEs in recent years is obtaining funding, which has become even more difficult due to the introduction of capital requirements. This has put pressure on banks' lending capacity. Basel III has added an additional burden by increasing the minimum capital ratio from 6% to 8%, which formed higher barriers for SMEs seeking financing. The significance of SME well-being on the worldwide economy was accentuated by different investigations. Applying an organization's reliability assessment is one of the key points a bank faces, prior to choosing, whether the organization ought to be endorsed for finance[4][5]. Restricted data accessibility, financial adaptability, and absence of lucidity are not many of many variables, which are confounding reliability assessments for SMEs, along these lines expanding credit chance and cost of supporting.

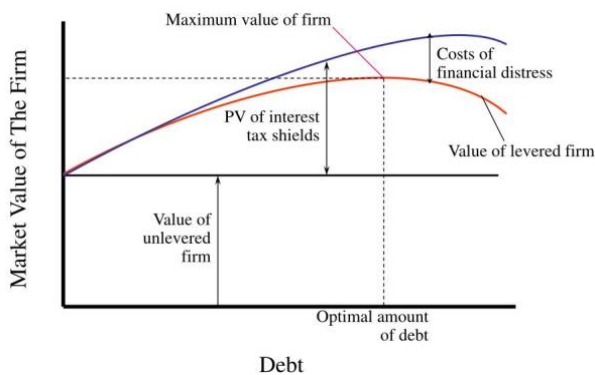

Figure 1: Financial Distress for Market Value and Debt

Financially distressed businesses are organizations that are encountering trouble in adapting to their everyday administration assignments. In the worst situation imaginable, these organizations are inclined to go to liquidation lengths (Baharin and Sentosa, 2013). On the off chance that a calculation can acquire a similar skill from previous data to work on the expectation for the future, it can bring about organizations distinguishing warnings adequately early, or possibly with perfect timing, to make a proper healing move. Studies regarding corporate disappointment forecast for the most part mean to decide one or a few factors that make it conceivable to segregate between falling flat and sound organizations (Refait, 2004). The expectation to a great extent centers around financial data. The trailblazers, Beaver (1966) and Altman (1968). utilized univariate and straight discriminant investigations individually[6][7]. Later examinations have utilized different factual methods like Logit and Probit models. Factual techniques, in any case, rely upon prohibitive theories. In the nineties, on account of the improvements in PC sciences, a few creators depended on Machine Learning calculations (ML) like brain organizations to foresee financial distress[8][9].

A decent financial distress expectation plot should be sensible and productive. Nonetheless, an enormous number of excess and inconsequential characteristics would influence characterization execution by expanding figuring costs and the time expected to learn and test the classifier. Highlight choice, as a significant innovation in data mining and machine learning, has been generally utilized in arrangement models. Choosing highlights before applying the characterization technique to the first dataset enjoys a few benefits, for example, refining the data, diminishing estimation cost, and further developing order exactness. Subsequently, we take on an element determination calculation to work on the nature of the financial distress forecast. In the field of financial distress forecast, numerous element determination techniques are proposed, like unpleasant set strategy, LASSO technique, covering, and channel[10][11]. Nonetheless, the vast majority of these methodologies neglect to consider the characteristics and data wellsprings of discrete elements and the various impacts they might have on the tag. the data qualities of an organization can be gathered by the examination of financial explanations and data sources, like the development capacity, dissolvability, working skill of financial proclamations and corporate administration qualities, and market exchange data. these alliances mirror the connection, overt repetitiveness, and complementarity between the elements. along these lines, it tends to be applied to resulting highlight choice strategies[12][13][14].

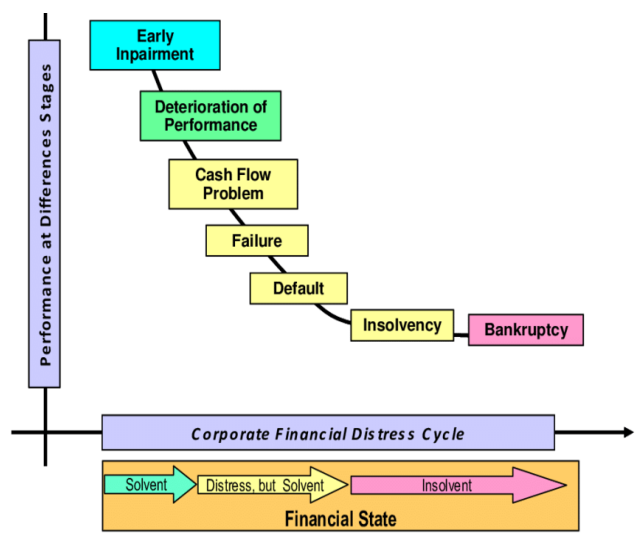

Figure 2: Process of Integral Financial Distress

While liquidation forecast mostly centers around foreseeing the finish of an organization's lifecycle, with the moderately slim likelihood of endurance through rebuilding, financial distress expectation is a more normal event, when occupational encounters impermanent issues conference its commitments. A few financial distresses could wind up as insolvencies however frequently they are simple difficulties and don't keep going long[15][16]. According to a bank point of view, while surveying reliability, it is vital to recognize borrowers which will actually want to meet their commitments without encountering installment hardships. A wide range of creators involved different methodologies and techniques for financial distress expectation and got various outcomes shifting from very high - 95% expectation achievement which was reached by Altman in his previous examinations, to 'near earth' indicators with results seething from 70 % to 85 %.

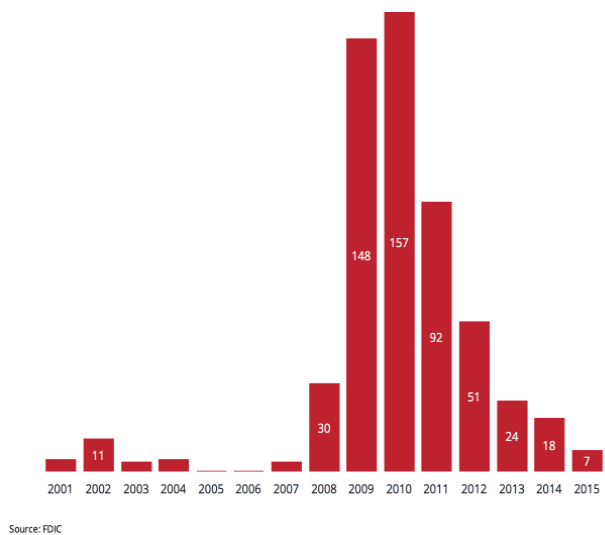

Figure 3: Financial Distress variation in Banks

High fluctuation in results, which have just hardly expanded all through the past 10 years, and ceaseless quest for a superior indicator demonstrates that beneficial devices for financial value assessment are required. Indeed, even though financial distress forecasting is a well-studied topic, there isn't much research on SME materials, particularly in terms of time factor inclusion and various machine learning process application. This paper's goal is to provide a financial distress anticipation model that incorporates time components and measures the perspective of the SME element's FICO score in a 1-year horizon by using the three most promising machine learning techniques. procedures. Despite past examinations, where the likelihood of default was displayed, this exploration assesses the FICO score viewpoint by demonstrating SME substance financial hardships, which won't be guaranteed to wind up as liquidations or defaults. Rather than utilizing factors that allow only a one-year financial proportion preview - extra time factors are added along with past-due history qualities, deals, and resource change markers[17][18][19].

The methodology taken for the ongoing review includes the use of Decision Trees, Artificial Neural Networks, and Naive Bayes, three Machine Learning techniques. These techniques will then be applied to predict a company's viability through the arrangement of several features from its bookkeeping. Unlike expectation models, which normally use crude proportions directly, the models developed in this study make use of the distinctions between determined proportions (their progression through periods). Additionally, the assessment is carried out on data sets that are not used for preparation to get rid of data overfitting. The goal of this study is to create a framework for financial forecasting that will be applied after being prepared using a variety of firms' verifiable final records. the created models to assess the idea of another organization's financial data. This paper regards this test as a characterization issue instead of a relapse issue. This implies that the real upsides of the last records won't be anticipated, yet rather endeavors to decide if an organization is dynamic or flopping in the accompanying financial period. [20][21]

## II. LITERATURE REVIEW

There are many purposes for the financial distress expectation models, including observing corporate dissolvability, evaluating the dangers of defaults on advances and securities, valuing credit subordinates, and different protections bearing credit chances. During the 1960s, two renowned papers were distributed that concerned the investigation of financial distress expectations and the development of forecast models. One is Beaver's concentrate in 1966 [22][23], where corporate financial distress was anticipated by univariate examination, and those with tremendous bank overdrafts, neglected profits on a favored stock, defaults on corporate securities, and announcements of chapter 11 were characterized as organizations in financial distress. In that review, organizations in financial distress from 1954 to 1964 were taken as tests, one more 79 ordinary organizations of comparable size and capital in a similar industry chose for coordinating, and 14 financial proportions, for example, the proportions estimating benefit, liquidity, and dissolvability, were chosen to anticipate the chance of corporate financial distress.

SMEs being inclined to higher inborn credit risk than bigger organizations, experience essentially stricter acknowledge value evaluation methods as well as higher insurance prerequisites. In this way, to guarantee lower obstructions to get the support it is critical to guarantee cutting-edge credit value evaluation models. In their examinations, the authors of such efficient research argued that experts who compare a new model to some reference classifier(s) have the propensity to favor the new model since model engineers are more skilled at their methodology and can tune it more lavishly than reference models. They found that the advantage of SMEs is strongly correlated with their size, whereas the correlation for large corporations was inverse. They made this observation by using discriminant and Logit techniques. While the effect was almost non-existent for large corporations, it was relative and not fixed for SME organizations. Fidrmuc and Hainz (2009) focused on the SME advance market in Slovakia using the Probit philosophy and discovered that the outcomes are questionable for the most part because of solid predispositions which are connected with industry explicitness as well as authoritative document impacts.

As per the outcomes, income/all-out liabilities were the most incredible in the financial distress forecast; in any case, the univariate model was simply ready to assess each factor in turn and couldn't think about other organization factors all in all. The other paper was Altman's concentrate in
1968 [24]. By working on the lack of a univariate model, a multivariate differential investigation model was laid out to foresee corporate financial distress. Those legitimately bankrupt, dominated, or perceived as rebuilt by the public chapter 11 regulation were characterized as organizations in financial distress. In that review, 33 organizations in financial distress from 1946 to 1965 were taken as tests, one more 33 financially sound organizations with comparative resources were chosen for coordinating, and 22 financial proportions were chosen from five significant extensive corporate aspects, like liquidity, benefit, financial influence, dissolvability, and turnover limit. Then, the proportions were consolidated to make an extensive list of the complete resources, X2: Retained profit/all-out resources, X3: Earnings before premium and duties/all-out resources, X4: Market worth of value/book worth of all-out obligation, and X5: Sales/all-out resources. In 2017, Altman et al. anticipated financial distress for organizations in 31 nations/locales (generally in Europe) by the Z-Score model, and its exactness was 70-80%[25][26][27].

Corporate disappointment expectation representations can be arranged in two: factual-based or calculation-determined utilizing Machine Learning (ML) methods. As referenced in the prologue to this paper, trailblazers of chapter 11 forecast utilized measurable strategies to segregate between fizzling and sound organizations (Refait, 2004). By the by, and albeit measurable strategies are as yet utilized, in the nineties, a few creators embraced Artificial Intelligence calculations (or Machine Learning) procedures like brain organizations for organization disappointment prediction[28][29].

Machine Learning (ML) is "the study of getting PCs to act starved of being customized". This cycle endeavors to recognize significant examples among the data sources and independently fabricate a prototype that can depict these examples without human intercession. ML instruments are worried about giving projects the capacity to learn and adjust to various examples. Following the strides of wise creatures, numerous abilities are gotten or refined through learning at various examples, rather than adhering to express guidelines. ML may likewise be characterized as the mind-boggling calculation cycle of programmed design acknowledgment and shrewd dynamic given preparation test data (Dua and Du, 2016). Subsequently, ML methods must be assessed exactly because their exhibition is vigorously subject to the preparation dataset. The elements chosen for displaying an issue assume a critical part of the expectation. There are numerous angles to consider while talking about financial distress. The forecast generally centers around economic data. Financial proportions can be named resource the executives' proportions, influence proportions (surveying the capacity of a firm to meet financial obligations), liquidity proportions (evaluating the capacity of a firm to meet its obligation obligations), and productivity proportions[30].

Huang et al. employed various brain network methodologies to assess the performance of different classifiers for credit risk assessment. They found that the Probabilistic Neural Network exhibited the highest discriminatory power and had the lowest overall and Type 2 error rates. Zhang et al. (2015) utilized state-of-the-art artificial intelligence techniques to incorporate Supply Chain Finance (SCF) into credit risk assessment. Similarly, Zhu et al. (2016) investigated the performance of Logistic Regression, Artificial Neural Networks, and their Hybrid models using Stress Concentration Factor as a feature. Their results showed that the Artificial Neural Networks had a lower Type 2 error rate compared to Logistic Regression, while the two-stage hybrid model achieved the best overall precision. To address the variability of financial efficacy assumptions in development due to linguistic representation, Shon et al. (2016) proposed the use of Fuzzy Logistic Regression. This approach was deemed appropriate to improve the accuracy of credit risk assessment compared to conventional Logistic Regression. Addo et al. (2018) constructed binary classifiers for credit risk assessment using AI techniques, including Logistic Regression, Random Forests, and Neural Networks. Model evaluation was based on factors from financial statements, and the top 10 features were selected and used to compare performance across different methods. The Tree-based classifiers were found to perform similarly and be more robust than multi-layer Artificial Neural Networks. Arora and Kaur (2020) utilized the Bollaso (Bootstrap-Lasso) feature selection method to identify relevant and stable variables from a pool of features. These variables were then used as inputs to various machine learning models, including Random Forest, Relevance Vector Machine, Naïve Bayes, and k-Nearest Neighbors, to evaluate their predictive accuracy for financial stability assessment. Consistent with previous studies, Random Forest exhibited the highest accuracy compared to other methods.

### III. METHODS AND METHODOLOGY

In the forecast of corporate financial distress, factors are isolated into a few gatherings as per market exchanges, development capacity, dissolvability, productivity, etc, and each gathering comprises a few factors. Right now, the univariate determination technique will overlook the data concealed in the variable social affair architecture, which could lessen the show of variable decisions and may even mis-select factors. there are more and more markers reflecting the monetary status of undertakings in reality, and countless of them are upheaval factors. Accepting all variables associated with the model capriciously, the precision of the model will be diminished. thusly, factors should be picked in the illustration. the advantage of pitiful head part assessment lies in its consistency under the little disrupting impact of information change and its penchant to vanquish multicollinearity ordinarily, and it can give an all-out factor choice way[31][32].

Relevance Vector Machine (RVM) is a standard machine learning characterization strategy now. Because of its benefits in tackling small examples and nonlinear issues and its great prescient presentation, it has been broadly applied practically speaking. thusly, joining the benefits of meager head part examination and Relevance Vector Machines, this paper proposes the GSPCA-RVM technique. Taking into account highlight gathering, the viability of meager head part examination in recognizing the main element pointers in every class is presented, which empowers us to construct a superior expectation model. Figure 1 is the flowchart of target acknowledgment joining SPCA and RVM. Furthermore, the particular strides of the calculation GSPCA-RVM are displayed as follows. Right off the bat, as per the data sources and financial proclamation investigation techniques, the trademark files of recorded organizations are partitioned into a few gatherings (like dissolvability, benefit, and development capacity). Second, utilize inadequate head part investigation to screen the traits of each social affair of documents. Third, join the brand name records screened by each social event into a new dataset, and conclude the readiness tests and test tests. then, at that point, input the planning tests with the RVM procedure, get the coefficient and deviation of the isolation capacity through learning, and foster the course of action model. Finally, input the test tests to the portrayal model, then, take the assumption handling, and lastly ascertain the precision[33][34].

GSPCA-RVM Algorithm. Considering that X addresses a normalized trademark lattice of n ∗ m, where n is the number of recorded organizations and m is the number of qualities of recorded organizations, the scanty head part examination is planned on the premise that the chief part investigation can be changed into a quadratic punishment relapse issue. that is, the arrangement of head parts is straightforwardly changed into LASSO relapse. in this way, the arrangement of meager head parts is successfully changed into the variable choice issue of the straight model. On this premise, the punishment construction of the versatile net is acquainted with acquiring the scanty head parts. the goal capability of inadequate head part examination is as per the following:

$$(\hat{a}, m\hat{\beta}) = argmin \sum_{i=1}^{n} \|X_i - \alpha\beta^M X_i\|^2 + \varepsilon\|\beta\|^2 \quad (1)$$

In this way, regression acquaintance is used to finding the initial principal component.

Along these lines, the first head part investigation is changed into a relapse issue. By adding LASSO punishment thing to the above condition, meager head parts can be gotten. in this way, the accompanying streamlining issue can be acquired:

$$(\hat{a}, m\hat{\beta}) = argmin \sum_{i=1}^{n} \|X_i - \alpha\beta^M X_i\|^2 + \varepsilon \sum_{k=1}^{n} \|\beta\|^2 + \sum_{k=1}^{n} \varepsilon_n \|\beta\|^2$$

As expressed over, the arrangement of meager head parts can be changed into a punishment relapse issue. the general LASSO punishment relapse issue can be settled by the least point relapse. subsequently, the computation of meager head parts can likewise be advantageously gotten by utilizing the least point relapse calculation.

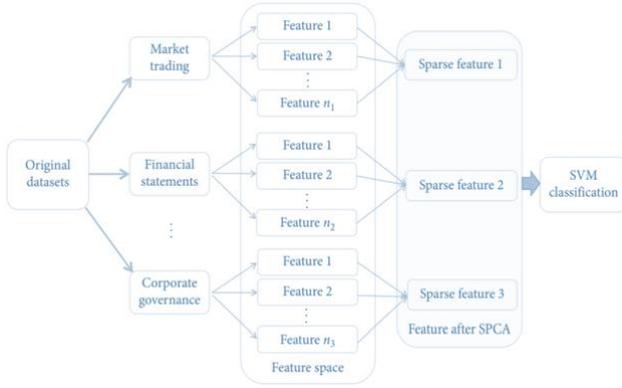

Figure 4: Flowchart of Combined SPCA and RVM

In the experimental exploration of financial distress expectation, scientists have not agreed on the choice of pointers. this paper endeavors to gather exhaustive data on financial qualities, exchange qualities and trade administration marks of China's A-share recorded organizations. It comprises economic elements, market exchange highlights, and corporate administration pointers. As per the order technique for financial files of recorded organizations in RESSET financial exploration With the expansion of market exchange attributes bunch and corporate administration qualities bunch, the dissemination of significant data and pointers reflected by each gathering.

*Principal Component Analysis*

As per the past segment, all attributes of recorded organizations are partitioned into 12 gatherings as per data sources and financial proclamation investigation techniques. Scanty head part Each data collection was examined, and the primary head portion coefficient of each collection was noted. It was selected if the trademark variable's major head part component was monotonic. It was eliminated because the coefficient was zero. The primary head component coefficient and selected trademark data for each group are displayed in Table 3. To determine whether collecting sparsity (GSPCA) improves the first dataset's essential data and kills repetitive data, we lead inadequate head part investigation for each of the 179 highlights (SPCA) and hold the trademark records with nonzero coefficients of the main four head parts.

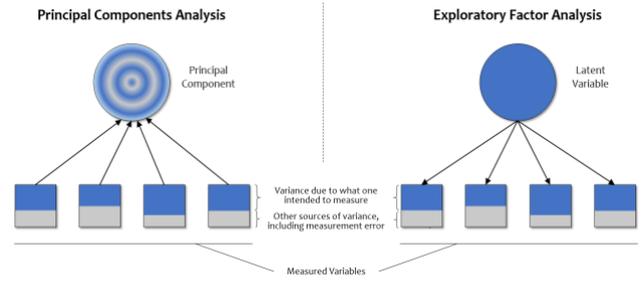

Figure 5: Graphical explanation of Principle Component Analysis

*Relevant Vector Machine*

The analytic tool of Relevant Vector Machine (RVM) uses a probabilistic erudition system to make expectations. The model gauges a likelihood , which is the likelihood of the class mark (significant burdensome issue) assumed novel information X. The RVM preparing methodology analyses back dissemination of loads, which depicts the 'significance' of each preparation model information to the prepared model. Other less significant preparing stage information is 'pruned' with their comparing loads set to nothing. RVM requires the determination of a portion of work and a bit width boundary. Here a non-straight Gaussian bit was utilized with the part width boundary decided utilizing the 'forget about one' cross-approval of the preparation informational index. In the ongoing review, the RVM calculation was executed utilizing a tool kit and in-house custom schedules as point by point somewhere else.

The characterization issue was addressed as

$\{X_n, t_n\}_{n=1}^N,$

where Xn addresses preparing stage information and tn-comparing objective marks which can either be consistent qualities (for example monetary trouble rating scores) for a relapse issue, or parallel grouping values for a characterization issue.

The above plan can be addressed as a standard direct model

$$t_n = y(X_n:W) + t_n \qquad (3)$$

Generalized to,

$$t_n = \sum_{n=1}^N \omega_n k(x, x_n) + b + \varepsilon_n \qquad (4)$$

Where $W = (\omega_1, \omega_2, \omega_3, \dots, \omega_N)^T$ is a weighting vector, b is a model predisposition recognized during the preparation cycle, Xn is an element vector and "n addresses estimation commotion (expected zero mean Gaussian circulation and difference). This is a non-direct Gaussian piece planning capability

$$k(x_1, x_2) = \exp(-\eta \|X_1 - X_2\|),$$

As examined in the fundamental text and Supplementary material, the piece width boundary was assessed utilizing a non-one-sided cross-approval system during the preparation stage. Highlight vectors with related weighting values were then utilized for making monetary pain expectations during the testing stage. The RVM calculation utilizes a Bayesian plan pointed toward assessing the back dispersion of weighting values:

Poster distribution = (LikelihoodthPrior)/marginal distribution.   (5)

The likelihood of a given dataset can be expressed as

$$\rho(t:W,\sigma^2) = (2\pi\sigma^2)^{-N/2} exp\left\{-\frac{1}{2\sigma^2}\|t-\phi W\|^2\right\} \quad (6)$$

A standard approach to avoid over-fitting (Tipping, 2001) is to introduce a zero-mean Gaussian prior distribution of the parameters defined as

$$\rho\left(\frac{W}{\alpha}\right) = \prod_{n=0}^{N} N(\omega_n: 0, \alpha_n^{-1}) \quad (7)$$

Where is a vector of N + 1 hyper-boundaries, each with a weighting limit coordinated with solitary hyper-boundaries. The way toward describing the priors of the hyperparameters (hyper-priors) and disturbance variance is depicted exhaustively elsewhere. At the point when the proof for the model was boosted as for the hyper-parameters, various hyper-parameters tend towards vastness compelling the comparing boundaries to nothing. These boundaries were pruned utilizing the programmed importance assurance technique (Mackay, 1992; Tipping, 2001) bringing about a scanty model.

Since the probability (4) and earlier (5) are characterized this permits the Bayesian equation (3) to be re-communicated as

$$p(W:t,\alpha,\sigma^2) = \frac{\rho(t:W,\sigma^2)p(W:\alpha)}{p(t:\alpha,\sigma^2)} \quad (8)$$

The boundaries α and σ were assessed utilizing a sort II greatest probability strategy. Forecasts made by weighting the premise capacities (4) by the back mean loads can be utilized for relapse issues. For characterization issues, as in the current investigation, a changed probability work was utilized. The goal was to foresee the back likelihood of enrolment of one of the two classes (significant burdensome issue versus controls) given preparing information. The direct model presented before the term was changed utilizing a calculated sigmoid capacity

$$\sigma(Y) = \frac{1}{(1+e-y)} toy(x,W) = \sigma\{W^T, \varphi(x)\}$$
$$= \frac{1}{1+exp\sigma\{W^T, \varphi(X)\}} \quad (9)$$

By adopting a Bernoulli distribution it is possible to derive:

$$p(t:w) = \prod_{n=1}^{N} \sigma\{y(x_n:W\}^t [1-\sigma\{y(x_n;W\}]^{1-t} \quad (10)$$

At long last, the back conveyance of the loads was assessed utilizing a Laplace technique. These loads were utilized for making RVM expectations.

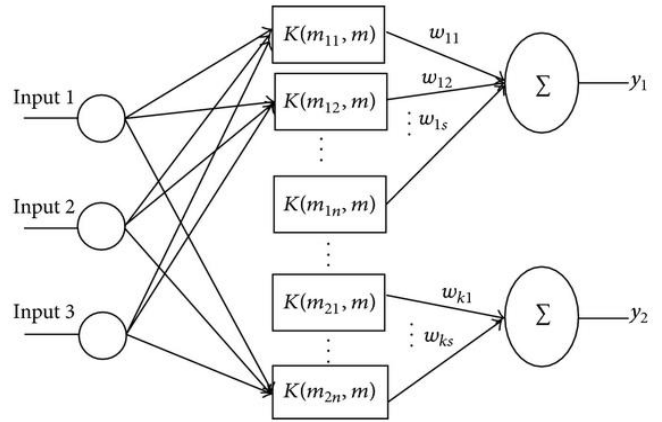

Figure 6: Relevance Vector Machine architecture

*Empirical Research on Proposed Method*

Figure outwardly mirrors the conveyance of all information elements of recorded organizations and the consequences of screening highlighted by inadequate head part investigation. From the gathering, the trademark marks (OF) mostly center around productivity, dissolvability, development capacity, capital design, and per-share pointers. through inadequate head part investigation of each gathering of highlights (GSPCA), a decent aspect decrease

impact can be accomplished. 61 highlights can be chosen from the first 179-pointers. Of the 40 markers in the element bunch mirroring the organization's productivity, just 3 were chosen by meager head part examination, and just 3 of the 21 pointers in the component bunch mirroring the list data per share were held. Generally, none of the signs of the trademark bunch mirroring the profit capacity of the organization or the trademark gathering of examination are inadequate, marks of the trademark bunch mirroring the development capacity of the organization are held. To analyze the impact of collection sparsity in highlight determination, this paper conducts a scanty head part examination on every one of the pointers (SPCA). this technique chooses 12 attributes from the first pointers, and these 12 markers chiefly mirror the data of the venture's capital construction, acquiring quality, and corporate administration. As can be finished from the Figure, albeit this technique accomplishes aspect decrease impact, it overlooks numerous parts of the organization like productivity, development capacity, activity capacity, and market exchange. It might eliminate a ton of helpful data connected with an organization's gauge of financial distress.

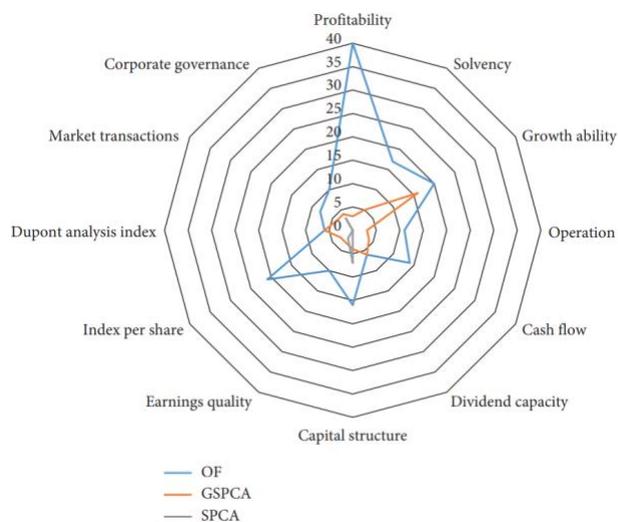

Figure 7: Feature results outcome by SPCA of empirical financial distress.

IV. DISCUSSION

To reflect genuine loan specialist credit dynamic structures and in opposition to the common likelihood of default indicators, subordinates are not set in stone by portraying monetary misery when an SME component faces monetary difficulties and does not show up at decisive periods of section 11. Customary monetary extents, proposed were used for picking free factors, which include a country identifier and monetary variables from monetary strength appraisal arrangements like Liquidity, Leverage, Profitability, Coverage, and Activity. To also additionally foster request execution - new free factors were consolidated. As confirmed by Shumway (2001), as opposed to using factors that permit only one-year monetary extent to see - additional year time factors were added alongside past due history, changes in assets, and arrangement characteristics. Disregarding Random Forests, the best-performing procedure for a static period gathering - Artificial Neural Networks - experienced disintegrating in assumption responsiveness by 6 p.p. Mixed classifier response to time components might be associated with the trial thought of model new development, which can move from variables to getting ready system contrasts (Addo et al., 2018; Arora and Kaur; 2020). By and large, multi-period request was a preferred philosophy in assessment over a static-period pointer, which shows that time factors diminished the weakness factor by updating for periods at serious risk. Individual variable importance, in the best performing Random Forests classifier with additional elements, was tantamount all through a larger piece of variables. Most vital importance was connected with association age and least to country marker. Modestly little importance numbers for individual variables show that the major classifier strength is connected with associations between all factors and not independent elements[35].

The assessed financial distress indicator, whenever contrasted with the traditional likelihood of default classifiers, had somewhat lower expectation power. In correlation, other likelihood of default displaying studies had exactness as high (even in the wake of rectifying for predisposition). Potential explanations behind lower assessor precision could be connected to various ward variable detailing. Old-style credit risk forecast models are centered around the reality of liquidation (or default), while in this study financial distress is looked at when an SME element starts having monetary issues, as determined by changes in FICO ratings. This part is a finishing occasion in the organization's presence, seldom happens at least a time or two and it takes more than one component to set off. Then again, organizations could encounter various financial distress occasions without setting off a chapter 11. For instance, occasions like court debates shut occupied capital offices, and staff changes can be purposes behind brief economic shocks, in any event, for organizations that recently demonstrated a low likelihood of default. According to the bank's point of view, contingent upon its gamble craving, neither one nor the other would be liked because of possibly higher capital necessities.

## V. Conclusion

In this paper, an RVM model in light of scanty head part examination (GSPCA-RVM) is proposed to manage financial distress expectations. Displaying credit risk is one of the significant subjects each moneylender faces, whether deciding the financial capacity of an insolvent to reimburse the credit or computing the principal prerequisites which are forced by the financial controller. This errand turns out to be significantly more enthusiastically when SME explicit variables like restricted data accessibility and financial adaptability are presented. In the component determination phase of the first dataset, we suggest a technique to bunch the highlights as per data causes and financial explanation examination. the reason for this strategy is to examine whether the prescient presentation of the model can be enhanced by choosing less, somewhat more significant factors from every data including classification. Contrasted and other estimating models, our strategy has a superior determining impact since it joins the administration technique and machine learning strategy in the field of big business financial distress gauging. Taking into account that the data element of recorded organizations has the attributes of regular gathering, applying the financial assertion examination strategy to the gathering of unique datasets keeps away from the normal aspect decrease technique overlooking the data concealed in the variable gathering structure, which might lessen the estimating impact of the model. Also, the meager calculation chooses less and more significant factors from every data highlight classification to further develop the expectation execution and logical capacity of the classical and extra examines which include classifications of the organization can give more data to foreseeing financial distress.